\documentclass[sigconf]{acmart}

\renewcommand\footnotetextcopyrightpermission[1]{} 

\AtBeginDocument{%
  }

\usepackage{booktabs}
\usepackage{array}
\usepackage{lipsum}
\usepackage{float}
\usepackage{makecell} % 用于单元格内换行

\usepackage{multirow} 
\usepackage{multicol} 
\usepackage{enumitem}
\usepackage{caption}

\begin{document}

\title{Text-Visual Semantic Constrained AI-Generated Image Quality Assessment}

\author{Qiang Li}
\email{mozhu87@mail.nwpu.edu.cn}
\affiliation{%
  \institution{Northwestern Polytechnical University}
  \city{Xi'an}
  \country{China}
}

\author{Qingsen Yan}
\email{yqs@mail.nwpu.edu.cn}
\affiliation{%
  \institution{Northwestern Polytechnical University}
  \city{Xi'an}
  \country{China}
}

\author{Haojian Huang}
\email{haojianhuang@connect.hku.hk}
\affiliation{%
  \institution{The University of Hong Kong}
  \city{Hong Kong}
  \country{China}
}

\author{Peng Wu}
\email{pengwu@nwpu.edu.cn}
\affiliation{%
  \institution{Northwestern Polytechnical University}
  \city{Xi'an}
  \country{China}
}

\author{Haokui Zhang}
\email{hkzhang@nwpu.edu.cn}
\affiliation{%
  \institution{Northwestern Polytechnical University}
  \city{Xi'an}
  \country{China}
}

\author{Yanning Zhang}
\email{zhangyanning@npu.edu.cn}
\affiliation{%
  \institution{Northwestern Polytechnical University}
  \city{Xi'an}
  \country{China}
}

% 设置短作者格式（用于页眉）
%\renewcommand{\shortauthors}{Li et al.}

\renewcommand{\shortauthors}{Qiang Li et al.}
\settopmatter{printacmref=false} %remove acm reference format

\begin{teaserfigure}
  \centering
  \includegraphics[width=\textwidth]{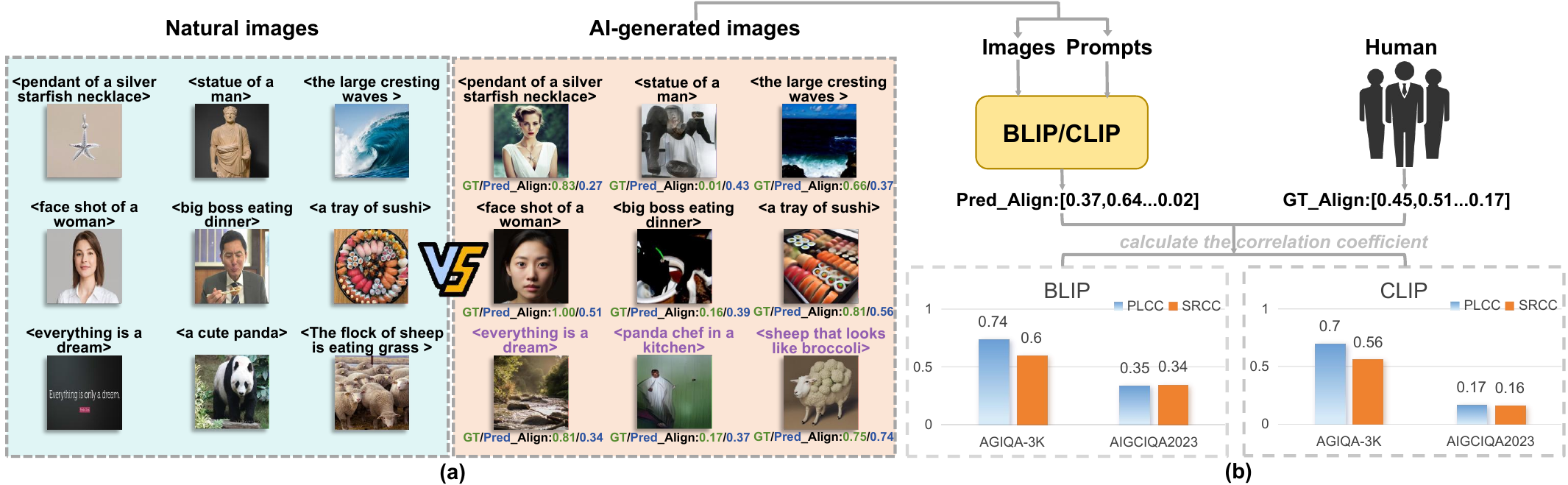}
  \caption{(a) Comparing natural images and AGIs; note AGIs can depict non-existent subjects (purple text examples). (b) Correlation (PLCC/SRCC) between predicted ($\mathrm{Pred}_{\mathrm{align}}$) and ground-truth ($\mathrm{GT}_{\mathrm{align}}$) image-text alignment scores.}
  \label{fig:fig0}
\end{teaserfigure}

\begin{abstract}
With the rapid advancements in Artificial Intelligence Generated Image (AGI) technology, the accurate assessment of their quality has become an increasingly vital requirement. Prevailing methods typically rely on cross-modal models like CLIP or BLIP to evaluate text-image alignment and visual quality. However, when applied to AGIs, these methods encounter two primary challenges: semantic misalignment and details perception missing. To address these limitations, we propose Text-Visual Semantic Constrained AI-Generated Image Quality Assessment (SC-AGIQA), a unified framework that leverages text-visual semantic constraints to significantly enhance the comprehensive evaluation of both text-image consistency and perceptual distortion in AI-generated images. Our approach integrates key capabilities from multiple models and tackles the aforementioned challenges by introducing two core modules: the Text-assisted Semantic Alignment Module (TSAM), which leverages Multimodal Large Language Models (MLLMs) to bridge the semantic gap by generating an image description and comparing it against the original prompt for a refined consistency check, and the Frequency-domain Fine-Grained Degradation Perception Module (FFDPM), which draws inspiration from Human Visual System (HVS) properties by employing frequency domain analysis combined with perceptual sensitivity weighting to better quantify subtle visual distortions and enhance the capture of fine-grained visual quality details in images. Extensive experiments conducted on multiple benchmark datasets demonstrate that SC-AGIQA outperforms existing state-of-the-art methods. The code is publicly available at https://github.com/mozhu1/SC-AGIQA.
\end{abstract}
%%https://anonymous.4open.science/r/SC-AGIQA-47C9





\maketitle
\pagestyle{plain}
\pagestyle{empty}

\vspace{-1em}
\section{Introduction}
With the rapid development of AI-Generated Images (AGIs) models, AI-Generated Image Quality Assessment (AGIQA) is emerging as an increasingly critical research area. The proliferation of advanced generative techniques, such as Diffusion Models~\cite{diffusion} and Generative Adversarial Networks (GANs)~\cite{gan}, produces vast amounts of imagery, yet their output quality varies significantly, underscoring the urgent need for reliable AGIQA methods. AGIQA aims to evaluate the quality of AI-synthesized images by emulating human subjective judgment, often considering both the image and its associated textual prompt. Distinct from traditional Image Quality Assessment (IQA), which primarily focuses on visual quality degradation from distortions like compression or noise, AGIQA addresses a broader and more complex evaluation spectrum. As highlighted by Tian et al.~\cite{agiqazongshu}, this necessitates evaluating various aspects, prioritizing text-image alignment and visual quality, alongside aesthetic appeal, content fairness, and toxicity.

While Blind Image Quality Assessment (BIQA) methods based on visual pre-trained models~\cite{zhang2020blind,su2020blindly,qin2023data,liu2023multiscale,chen2024topiq,agnolucci2024arniqa} excel at assessing perceptual visual quality, their inherent inability to process textual prompts renders them unsuitable for AGIQA tasks. Consequently, approaches leveraging cross-modal pre-trained models like CLIP~\cite{clip} or BLIP~\cite{blip} have gained prominence, favored for the inherent text-image alignment capabilities acquired during their pre-training. These methods typically assess AGI quality by evaluating the consistency between the generated image and associated textual information, whether it's the original prompt or quality-specific descriptions~\cite{IPCE, CLIPAGIQA, SFIQA, BTPAGIQA, TSP-MGS}. Others focus on modeling human preferences learned from large-scale datasets~\cite{xu2023imagereward, pick}. Such cross-modal strategies have demonstrated substantial performance gains over vision-only BIQA methods when evaluated on AGIQA datasets, with recent works achieving state-of-the-art results by integrating different aspects of quality assessment~\cite{MoE-AGIQA}.

Despite these advancements, methods based on cross-modal models still face two primary challenges:(1) Semantic Misalignment: The distribution gap between training data and AGI outputs, combined with the unique characteristics of AI-generated images, lies at the core of this issue. Cross-modal models like CLIP~\cite{clip} and BLIP~\cite{blip} are typically pre-trained on large-scale datasets of natural images (e.g., datasets like LAION~\cite{laion}, COCO~\cite{coco}, etc.), which exhibit significant distributional differences compared to AGI outputs (Figure~\ref{fig:fig0}(a)). AGIs often generate images with unconventional features, such as unusual lighting or object morphologies. Experiments on AGIQA-3K~\cite{agiqa3k} and AIGCIQA2023~\cite{aigciqa2023} (Figure~\ref{fig:fig0}(b)) revealed semantic misalignment in cross-modal models: the advanced BLIP~\cite{blip} model achieved an SRCC of 0.6 on AGIQA-3K and only 0.35 on the more challenging AIGCIQA2023 dataset, highlighting significant challenges in perceiving AGI semantics. (2) Lack of Detail Perception: Cross-modal pre-training strategies predominantly focus on optimizing instance-level semantic correspondence between text and vision. The lack of explicit modeling and enhancement of region-level low-level visual features weakens the models' ability to detect and quantify fine-grained distortions. This significantly hinders their accuracy in visual quality assessment.

To address these challenges, we propose two key modules: the Text-assisted Semantic Alignment Module and the Frequency-domain Fine-Grained Degradation Perception Module. The former leverages the advanced understanding capabilities of Multimodal Large Language Models (MLLMs) for AGIs. It translates key aspects of the generated image, encompassing both content and aesthetics, into a textual "descriptive prompt". This shifts the semantic consistency assessment: instead of directly comparing the AGI image with the original prompt, it compares the descriptive prompt with the original prompt. By leveraging the MLLM's robust comprehension, this module aims to mitigate the semantic misalignment issues encountered by previous approaches. The latter module, built upon visual pre-trained models, incorporates priors from the Human Visual System (HVS), specifically its varying sensitivity to spatial frequencies in distortion perception. This design enhances the capture and quantification of fine-grained visual details and artifacts, addressing the lack of detail perception. Finally, the information regarding semantic consistency and visual quality is effectively integrated using a Mixture-of-Experts Regression (MoER) framework to predict the overall quality score. Our contributions are summarized as follows:
%\vspace{-1.5em}
\begin{itemize}[leftmargin=*]
\item We propose a novel Text-assisted Semantic Alignment Module leveraging MLLM semantic understanding to mitigate semantic misalignment in AGI evaluation by comparing original against MLLM-generated descriptive prompts.
\item We design an HVS-inspired Frequency-domain Fine-Grained Degradation Perception Module incorporating frequency analysis to enhance perception and quantification of visual distortions and details, addressing AGIQA detail perception limitations.
\item We introduce SC-AGIQA, a unified framework effectively synergizing MLLM semantic understanding, cross-modal models for multimodal encoding, and specialized visual models for fine-grained distortion perception. This integrated approach achieves new state-of-the-art performance on challenging AGIQA benchmarks, namely AGIQA-1K, AGIQA-3K, and AIGCIQA2023.
\end{itemize}
\vspace{-1em}
\section{Related Work}
\label{sec:related_work} % Optional: Add a label for cross-referencing

\subsection{Blind Image Quality Assessment}
\label{subsec:traditional_iqa} % Optional: Label for the subsection

BIQA aims to build regression models that predict the perceptual quality score of distorted images without access to the original reference image, thereby simulating the human visual system's preference judgments on image quality. Early BIQA research primarily utilized Natural Scene Statistics methods \cite{niqe}. These methods are based on the assumption that natural images possess inherent statistical regularities and that distortions disrupt these regularities. They assess image quality by observing image statistical features from the spatial domain \cite{mittal2012no}, frequency domain \cite{moorthy2011blind,zhang2015feature}, and quantifying the degree of deviation from natural statistical models.

The rise of deep learning, particularly the success of visual models pre-trained on the large-scale ImageNet \cite{imagenet1k} dataset, has promoted the development of deep learning-based methods, bringing significant changes to the BIQA field and significantly improving evaluation performance. Researchers began to utilize models based on pre-trained Convolutional Neural Networks \cite{zhang2020blind,su2020blindly,liu2023multiscale,agnolucci2024arniqa} or Transformers \cite{qin2023data}, or hybrid models \cite{chen2024topiq} combining both to simultaneously capture local details and long-range dependencies important for assessing image quality, extracting quality-aware features and regressing scores. In recent years, with the development of powerful image-text alignment models such as CLIP \cite{clip}, BLIP \cite{blip}, research in the BIQA field has also been inspired. Researchers have begun experimenting with these models \cite{wang2023exploring,zhang2023blind}, often introducing quality-related text prompts or descriptions to guide the quality assessment process, leveraging the rich semantic understanding embedded within these pre-trained models. Following this trend, the rapid development of MLLMs has opened up new possibilities for BIQA. Researchers \cite{zhang2024qbench,wu2024qinstruct,wu2024qinstruct} are exploring the application of MLLMs to BIQA, not only because they can regress quality scores, but also because they possess the unique capability to generate comprehensive, descriptive text evaluating image quality from multiple aspects, providing richer evaluation information and better interpretability.

However, these emerging methods based on image-text alignment models and MLLMs currently still face a significant challenge. Their pre-training task objectives are not perfectly aligned with the core fine-grained perceptual quality score regression task. Consequently, although these models excel in semantic understanding and generating descriptive evaluations, in terms of accuracy for predicting numerical quality scores, due to the misalignment between their pre-training tasks and low-level vision tasks, they currently tend to lag behind those deep learning models specifically designed and optimized for the BIQA task.

\vspace{-1em}
\subsection{AI-generated Image Quality Assessment}
\label{subsec:aigc_iqa}
With the advancements in text-to-image synthesis technology, evaluating the quality of AGIs has become increasingly important. Unlike traditional BIQA, AGIQA requires evaluating images considering both the text prompt and the image content. Its evaluation dimensions are also more complex; Tian et al.~\cite{agiqazongshu} identify text-image consistency and visual quality as the most critical aspects. Traditional BIQA methods, unable to process text, are unsuitable for AGIQA, leading to the adoption of methods based on cross-modal models like CLIP \cite{clip} and BLIP \cite{blip}.

To facilitate research, several specialized datasets using the Mean Opinion Score (MOS) have been created, such as AGIQA-1k \cite{agiqa1k}, AGIQA-3k \cite{agiqa3k}, AIGCIQA2023 \cite{aigciqa2023}, and AIGCIQA-20k \cite{agiqa20k}. Building on cross-modal foundations and these datasets, various AGIQA approaches have emerged. Methods like TIER~\cite{TIER}, IPCE~\cite{IPCE}, CLIP-AGIQA~\cite{CLIPAGIQA}, SF-IQA~\cite{SFIQA}, IP-IQA~\cite{BTPAGIQA}, and TSP-MGS~\cite{TSP-MGS} typically assess image quality by evaluating the alignment between the image and either the original generation prompt or specific quality-related textual descriptions. For instance, IPCE~\cite{IPCE} pioneered this by constructing textual templates (e.g., "A photo that {adv} matches {prompt}") and utilizing CLIP to measure the matching degree between these templates and the image to regress a quality score. Such cross-modal approaches have demonstrated substantial performance gains over vision-only BIQA methods when evaluated on AGIQA datasets. In contrast, MA-AGIQA~\cite{MA-AGIQA} aids in quality assessment by analyzing whether AGIs contain coherent semantics using MLLM. Distinct from direct score regression, methods like ImageReward~\cite{xu2023imagereward} and PickScore~\cite{pick} focus on modeling human preferences. They collect large datasets of text-prompt/image pairs online and fine-tune cross-modal models using reward modeling techniques, thereby creating preference-aware pre-trained models. Based on this preference modeling foundation, MoE-AGIQA~\cite{MoE-AGIQA} recently integrated a distortion-aware branch with the ImageReward~\cite{xu2023imagereward} framework, concurrently modeling semantic information and distortion artifacts using a Mixture-of-Experts architecture to achieve new state-of-the-art results.

Nevertheless, AGIQA methods relying solely on these cross-modal models still face significant bottlenecks. These include challenges such as semantic misalignment, partly due to the distributional differences between their pre-training data (often natural images) and AGIs, and missing detail perception, as pre-training objectives frequently lack emphasis on the low-level visual features vital for accurate quality assessment.

\begin{figure*}[ht]
  \centering
  \includegraphics[width=\textwidth]{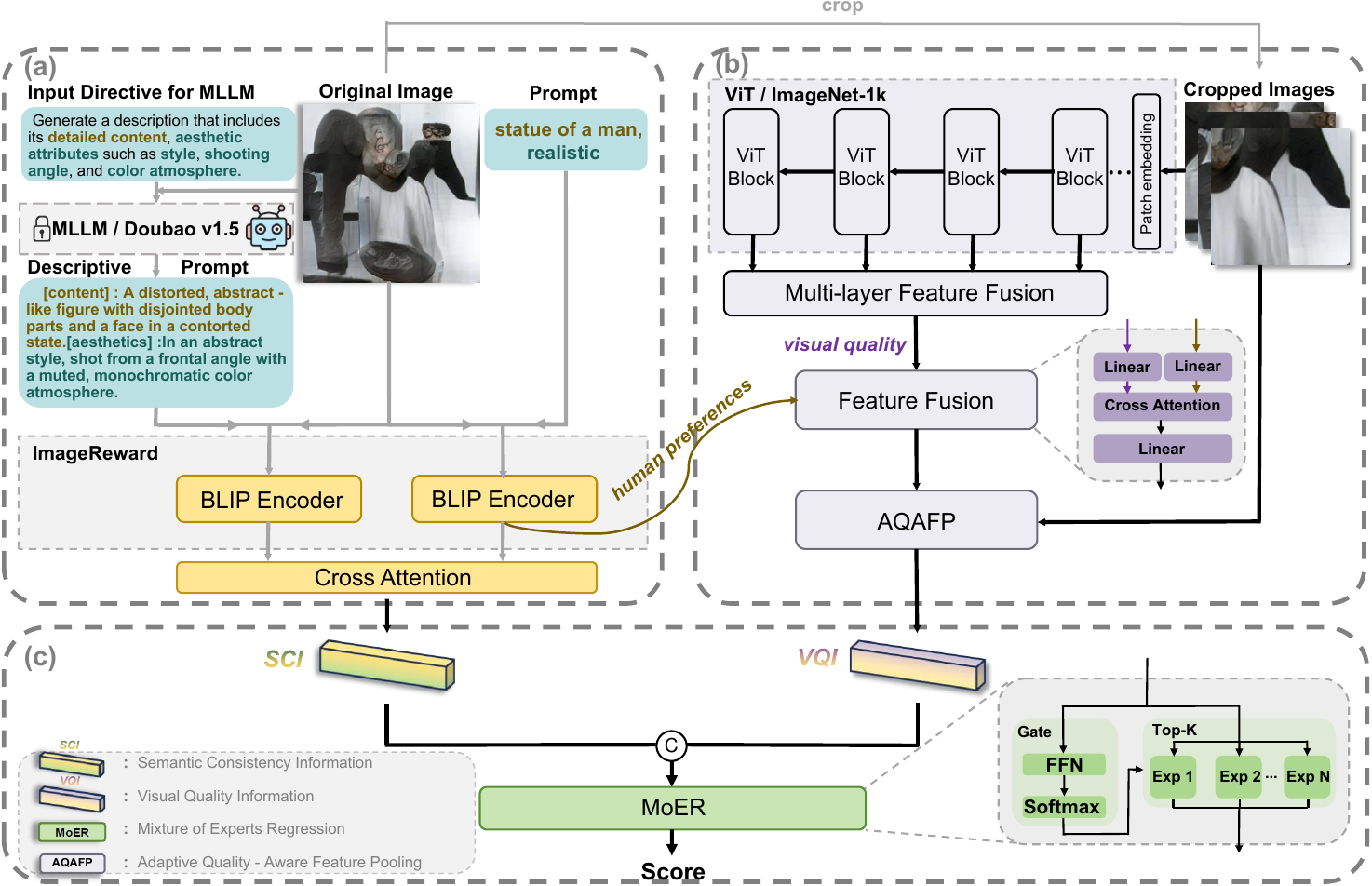}
  \vspace{-1em}
  \caption{Overview of the proposed SC-AGIQA framework. 
    (A) The Text-assisted Semantic Alignment Module (TSAM) leverages an MLLM to generate a descriptive prompt ($P_d$) from the image ($I_o$). Semantic Consistency Information ($SCI$) is obtained by comparing $P_d$ with the original prompt ($P_o$) via encoded features and cross-attention. 
    (B) The Frequency-domain Fine-grained Degradation Perception Module (FFDPM) extracts multi-level visual features from cropped images ($I_c$) using a ViT backbone. These features incorporate human preference signals (following~\cite{MoE-AGIQA}) before our proposed Adaptive Quality-Aware Feature Pooling (AQAFP), which integrates HVS priors via frequency analysis, produces the final Visual Quality Information ($VQI$). 
    (C) Finally, a Mixture-of-Experts (MoER) head integrates the $SCI$ and $VQI$ representations to predict the overall quality Score.}
  \label{fig:fig1}
  \vspace{-12pt}
\end{figure*}

\section{Method}
\label{sec:method}

% Our proposed SC-AGIQA framework, depicted in Figure~\ref{fig:fig1}, is designed to comprehensively assess AI-Generated Image (AGI) quality by tackling the prevalent challenges of semantic misalignment and inadequate detail perception. It comprises two specialized modules: the Text-assisted Semantic Alignment Module (TSAM) focusing on text-image consistency, and the Frequency-domain Fine-grained Degradation Perception Module (FFDPM) targeting visual quality. The outputs of these modules are integrated via a Mixture-of-Experts (MoER) regressor to yield the final quality score.
\subsection{Overview}
\label{ssec:overview}

The objective of AGIQA is to predict a quality score ($S_o$) reflecting human judgment for a generated image ($I_o$) conditioned on its input prompt ($P_o$). Formally, we aim to learn a function $f$:
\begin{equation}
\label{eq:agiqa_task_concise}
S_o = f(P_o, I_o)
\end{equation}
Current approaches often falter due to two primary challenges: accurately evaluating semantic alignment between the nuanced prompt $P_o$ and the generated content $I_o$, and precisely quantifying fine-grained visual degradations unique to AGIs that impact perceived quality but are missed by standard metrics.

To address these limitations, we propose the SC-AGIQA framework (Figure~\ref{fig:fig1}). SC-AGIQA decomposes the assessment into specialized modules targeting these distinct challenges. The Text-assisted Semantic Alignment Module (TSAM), processing the original prompt ($P_o$) and image ($I_o$), leverages an MLLM to perform a \textit{more robust semantic consistency check}, mitigating alignment issues and outputting a \textit{Semantic Consistency Information} ($SCI$) vector. Concurrently, the Frequency-domain Fine-grained Degradation Perception Module (FFDPM), analyzing image patches ($I_c$) derived from $I_o$, incorporates Human Visual System (HVS) principles via frequency analysis to \textit{enhance sensitivity to subtle visual artifacts}, yielding a \textit{Visual Quality Information} ($VQI$) vector. Finally, a Mixture-of-Experts (MoER) regressor takes the resulting $SCI$ and $VQI$ vectors as input, intelligently integrating this complementary information capturing both semantic alignment and perceptual detail to produce the final quality score $S_o$. The following sections detail each component.
\subsection{Text-assisted Semantic Alignment Module (TSAM)}
\label{ssec:tsam}

To accurately gauge the semantic consistency between the generated image $I_o$ and the original prompt $P_o$, especially given the unique characteristics of AGIs that challenge standard cross-modal models, the TSAM employs a MLLM, specifically Doubao v1.5 as illustrated in Figure~\ref{fig:fig1}(a). Instead of direct $P_o$-$I_o$ comparison, we first generate a descriptive prompt $P_d$ using the MLLM, guided by a specific directive to detail the image's content and aesthetics (as user prompts frequently specify aesthetic attributes such as style, angle etc.):
\begin{equation}
P_d = \text{MLLM}(\text{Directive}, I_o).
\label{eq:mllm_pd}
\end{equation}
This $P_d$ provides an MLLM-validated textual representation of the image.

We then leverage a pre-trained BLIP Encoder~\cite{blip}, potentially enhanced with human preference learning via frameworks like ImageReward~\cite{xu2023imagereward}, to embed both the descriptive prompt ($P_d$) and the original prompt ($P_o$) into a shared multimodal space, conditioned on the original image $I_o$:
\begin{equation}
\begin{aligned}
F_{P_dI_o} &=\text{BLIP-Encoder}(P_d, I_o) \\
F_{P_oI_o} &=\text{BLIP-Encoder}(P_o, I_o).
\end{aligned}
\label{eq:blip_enc}
\end{equation}
Here, $F_{P_dI_o}, F_{P_oI_o} \in \mathbb{R}^{N \times D}$ represent the multimodal feature sequences. The feature $F_{P_oI_o}$, derived from the original prompt and image using the potentially preference-aware encoder, also implicitly captures information related to human preferences.

The core idea is that the semantic alignment between $P_o$ and $P_d$, both grounded in $I_o$, reflects the true text-image consistency. We compute this alignment by first applying a cross-attention mechanism, where $F_{P_dI_o}$ acts as the query and $F_{P_oI_o}$ serves as the key and value. The resulting attention output features, which capture the token-level interactions, are then immediately aggregated using mean pooling along the token dimension ($N$) to produce the final fixed-size Semantic Consistency Information ($SCI$) vector:
\begin{equation}
SCI = \text{MeanPool}_N\left( \text{CrossAttention}(F_{P_dI_o}, F_{P_oI_o}, F_{P_oI_o}) \right).
\label{eq:cross_attn_pooled}
\end{equation}
This yields $SCI \in \mathbb{R}^{D}$, a compact representation of the overall semantic consistency ready for integration with the visual quality features.
\subsection{Frequency-domain Fine-grained Degradation Perception Module (FFDPM)}

To enhance the model's sensitivity to subtle visual artifacts and fine-grained details often missed by conventional methods, the FFDPM operates on cropped image patches $I_c$ derived from $I_o$, as shown in Figure~\ref{fig:fig1}(b). It utilizes a Vision Transformer (ViT)~\cite{vit}, pre-trained on ImageNet-1k \cite{imagenet1k}, as the backbone for extracting rich visual features. We exploit features from multiple deeper layers of the ViT, which capture complex patterns relevant to quality degradation:
\begin{equation}
[F_{v_1}, F_{v_2}, \dots, F_{v_L}] = \text{ViT}(I_c).
\label{eq:vit_features}
\end{equation}
\label{ssec:ffdpm}
\begin{figure}[ht] % [ht] 表示尽量在当前位置（here）或顶部（top）放置图片
    \centering % 图片居中显示
    \includegraphics[width=\columnwidth]{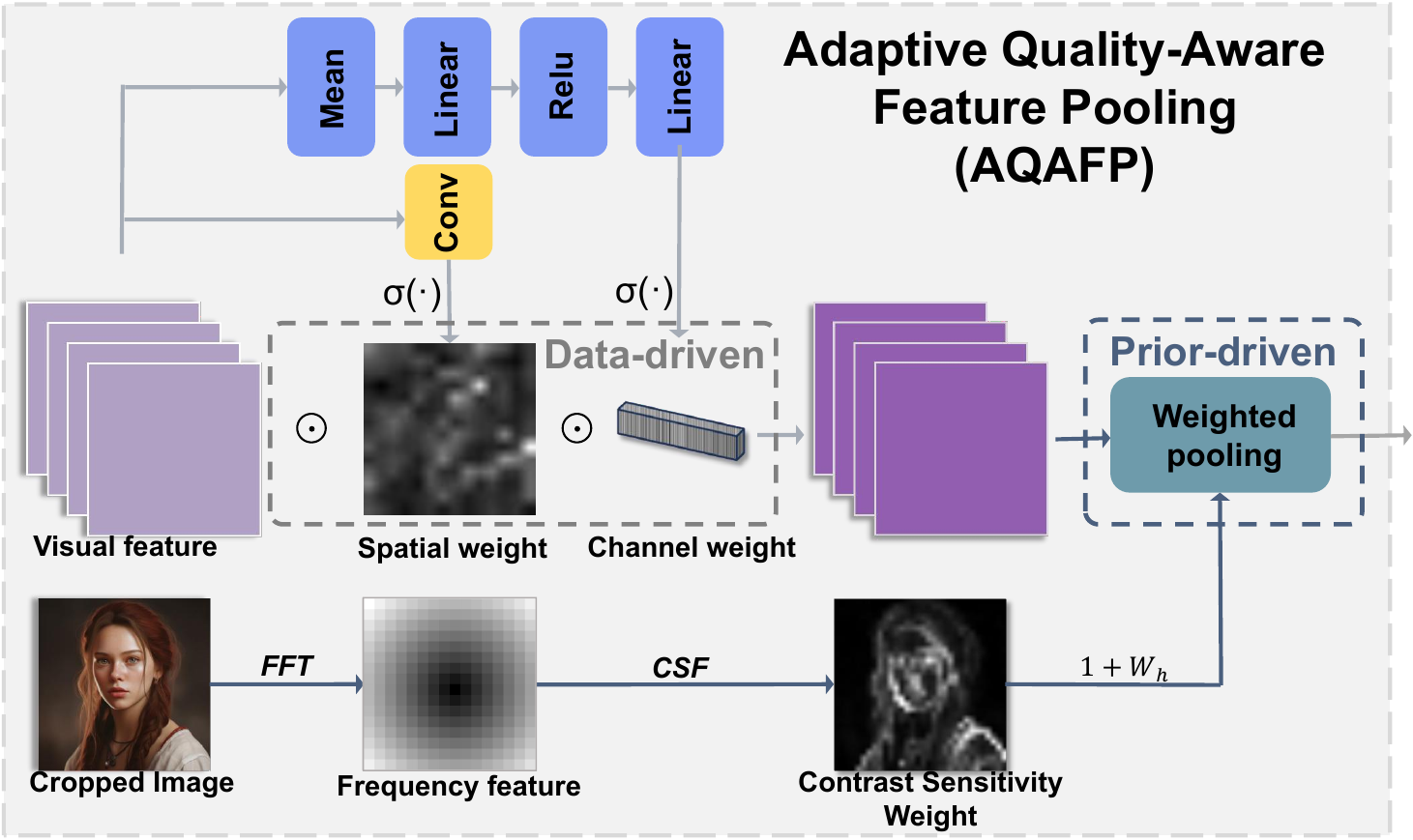} % 插入图片，宽度为单栏宽度
    \vspace{-1em}
    \caption{The principle of the Adaptive Quality-Aware Feature Pooling Module} % 图片标题
    \label{fig:fig2} % 图片标签，用于交叉引用
    \vspace{-2em}
\end{figure}
Features from the later four blocks ($F_{v_9}$ to $F_{v_{12}}$) are concatenated and passed through a Multi-layer Feature Fusion block (e.g., using convolutions) to obtain an intermediate visual quality representation $F_{vq} \in \mathbb{R}^{P \times D}$, where $P$ is the number of patches.

To further enhance the visual quality representation, we adopt the strategy proposed in MoE-AGIQA~\cite{MoE-AGIQA} for incorporating human preference signals, as visualized in Figure~\ref{fig:fig1}(b). The visual quality features $F_{vq}$, derived from the ViT backbone, are processed through a dedicated "Feature Fusion" block. Within this block, $F_{vq}$ is integrated with the preference-related features $F_{P_oI_o}$ obtained from the BLIP Encoder (Eq.~\ref{eq:blip_enc}). Following the MoE-AGIQA approach, this fusion can be implemented using mechanisms such as cross-attention between linearly projected versions of $F_{vq}$ and $F_{P_oI_o}$, followed by subsequent linear transformations. This step yields enhanced features, denoted as $F'_{vq}$, which effectively embed both intrinsic visual quality characteristics captured by the ViT and the preference-aware semantics inherent in $F_{P_oI_o}$. These fused features $F'_{vq} \in \mathbb{R}^{P \times D}$  then serve as the input to the pooling stage.

The final step in the FFDPM generates the Visual Quality Information ($VQI$) vector using our proposed Adaptive Quality-Aware Feature Pooling (AQAFP) module (Figure~\ref{fig:fig2}). Operating on the fused features $F'_{vq}$, the AQAFP implements an adaptive pooling strategy informed by both data-driven importance and Human Visual System (HVS) priors. First, it computes data-driven spatial ($W_s$) and channel ($W_c$) importance weights to emphasize quality-relevant feature dimensions and locations within $F'_{vq}$:
\begin{align}
    W_s &= \sigma(\text{Conv}_{1\times1}(F'_{vq})) \label{eq:spatial_weight_adapted}\\
    W_c &= \sigma(\text{Linear}(\text{MeanPool}_P(F'_{vq}))). \label{eq:channel_weight_adapted}
\end{align}
Here, $\sigma$ denotes the sigmoid function, $\text{Conv}_{1\times1}$ is a $1 \times 1$ convolution, and $\text{MeanPool}_P$ indicates mean pooling across the patch dimension~$P$.

Concurrently, to incorporate HVS priors, we calculate frequency-based importance weights ($W_h$). This process begins with the original cropped image patches $I_c$ corresponding spatially to the feature patches in $F'_{vq}$. Each image patch undergoes a Fast Fourier Transform (FFT) to obtain its frequency representation. We then apply a Contrast Sensitivity Function (CSF)~\cite{csf}, specifically formulated as $A(f) \approx 2.6(0.0192 + 0.114 f)e^{-(0.114 f)^{1.1}}$ where $f$ is the spatial frequency, to the frequency magnitudes, yielding a sensitivity score for each patch based on its dominant frequency content's perceptual relevance. These scores are passed through a sigmoid function $\sigma$ to normalize them into weights $W_h \in \mathbb{R}^{P \times 1}$:
\begin{equation}
    W_h = \sigma(\text{CSF}(\text{FFT}(I_{c, \text{patches}}))).
    \label{eq:hvs_weights}
\end{equation}
Finally, the AQAFP module produces the compact $VQI$ vector by performing weighted pooling on the features $F'_{vq}$. The features are first modulated by the learned spatial and channel importance ($F'_{vq} \odot W_s \odot W_c$) and then pooled across the patch dimension $P$, where the contribution of each patch is explicitly weighted by $(1 + W_h)$ to integrate the HVS prior:
\begin{equation}
    VQI = \text{WeightedPool}_{P}\left(F'_{vq} \odot W_s \odot W_c, 1 + W_h \right).
\label{eq:aqafp_weighted_pool_final}
\end{equation}
Here, $\odot$ denotes element-wise multiplication with broadcasting. This ensures that patches containing perceptually salient frequencies and features highlighted by the data-driven weights contribute more significantly to the final $VQI \in \mathbb{R}^{D}$ representation.

\subsection{Mixture-of-Experts Regression}
\label{ssec:moer}

As shown in Figure~\ref{fig:fig1}(c), the complementary Semantic Consistency Information ($SCI$) vector from TSAM and the Visual Quality Information ($VQI$) vector from FFDPM are concatenated to form a comprehensive representation of the AGI's quality aspects. This combined feature vector is fed into a Mixture-of-Experts Regression (MoER) head for final score prediction. The MoER architecture mimics the multi-expert mechanism of subjective evaluation. It consists of four parallel expert networks and a gating network (FFN + Softmax) that dynamically selects the Top-3 experts for weighted prediction based on input features. This allows the model to potentially learn specialized functions for different quality regimes or feature combinations:
\begin{equation}
S_o = \text{MoER}(\text{Concat}[SCI, VQI]).
\label{eq:moe_reg_final}
\end{equation}

The entire SC-AGIQA model is trained end-to-end by minimizing the Smooth L1 loss ($\mathcal{L}_{SmoothL1}$) between the predicted score $S_o$ and the ground-truth human score $S_{gt}$, encouraging accurate and robust quality prediction:
\begin{equation}
\mathcal{L}_{SmoothL1} = \sum_{i} \begin{cases}
0.5 (S_{o,i} - S_{gt,i})^2, & \text{if } |S_{o,i} - S_{gt,i}| < \beta \\
|S_{o,i} - S_{gt,i}| - 0.5 \beta, & \text{otherwise,}
\end{cases}
\label{eq:smoothl1_final}
\end{equation}
with $\beta$ typically set to 1.0.

\section{Experiment}
\label{sec:experiment}

\subsection{Datasets and Evaluation Metrics}
\label{subsec:datasets_metrics}

\textbf{Datasets.} We evaluate our method on three AGIQA benchmarks: AGIQA-1k~\cite{agiqa1k}, AGIQA-3k~\cite{agiqa3k}, and AIGCIQA2023~\cite{aigciqa2023}. AGIQA-1k~\cite{agiqa1k} provides 1,080 images from two early models. AGIQA-3k~\cite{agiqa3k} expands on this with 2,982 images from six diverse generation models (diffusion, GAN, auto-regressive) and includes MOS annotations for quality and text correspondence. AIGCIQA2023~\cite{aigciqa2023} offers 2,400 images from six recent models, featuring MOS ratings for quality, correspondence, and authenticity. These datasets collectively provide a comprehensive testbed with varied image content, generation techniques, and human quality judgments. For all experiments, we randomly partition each dataset, using 80\% for training and 20\% for validation, ensuring reproducibility via a fixed random seed.

\textbf{Evaluation Metrics.} To quantitatively assess performance, we employ the Spearman Rank-order Correlation Coefficient (SRCC) and the Pearson Linear Correlation Coefficient (PLCC), comparing our model's predicted scores ($y$) against the ground-truth MOS ($x$). SRCC measures prediction monotonicity, focusing on rank correlation:
\begin{equation}
\text{SRCC} = 1 - \frac{6 \sum_{i=1}^{n} d_i^2}{n(n^2 - 1)},
\label{eq:srcc}
\end{equation}
where $d_i$ is the rank difference for sample $i$, and $n$ is the sample count. PLCC assesses linear correlation and prediction accuracy:
\begin{equation}
\text{PLCC} = \frac{\text{cov}(x, y)}{\sigma_x \sigma_y},
\label{eq:plcc}
\end{equation}
where $\text{cov}(x, y)$ is the covariance, and $\sigma_x, \sigma_y$ are standard deviations.

Additionally, we introduce a 'Main Score' as a consolidated performance indicator, facilitating overall comparison across different methods and datasets. It is calculated as the arithmetic mean of SRCC and PLCC: \begin{equation} \text{Main Score} = \frac{\text{SRCC} + \text{PLCC}}{2}. \label{eq:main_score} \end{equation} For all three metrics (SRCC, PLCC, and Main Score), values closer to 1 indicate better alignment with human judgments and higher overall performance.

\subsection{Implementation Details}
% Model Components and Pre-training
\textbf{Model Components and Pre-training.} As illustrated in Figure~\ref{fig:fig1}(a), we employ the Doubao v1.5 MLLM (\texttt{doubao-1.5-vision-
pro-32k-250115} checkpoint) and a BLIP image encoder initialized with ImageReward~\cite{xu2023imagereward} weights. The component shown in Figure~\ref{fig:fig1}(b) is a Vision Transformer (ViT)~\cite{vit} backbone, pre-trained on ImageNet-1k~\cite{imagenet1k}. All multi-head cross-attention layers utilize 8 attention heads.

% Training Hyperparameters and Procedure
\textbf{Training Hyperparameters and Procedure.} We train the model end-to-end using the AdamW optimizer with parameters $\beta_1=0.9$, $\beta_2=0.999$, $\epsilon=1 \times 10^{-8}$, and a weight decay of 0.05. The learning rate schedule begins with a base learning rate of $1.0 \times 10^{-5}$, preceded by a 3-epoch linear warmup phase starting from $2.0 \times 10^{-6}$. Following the warmup, we employ a step learning rate decay strategy, reducing the learning rate by a factor of 0.1 every 3 epochs. We set the maximum number of training epochs to 100, utilizing an early stopping mechanism based on validation set performance to mitigate overfitting and determine the final model checkpoint. The batch size is configured based on the dataset: 12 for AGIQA-1k~\cite{agiqa1k} and AIGCIQA2023~\cite{aigciqa2023}, and 16 for the larger AGIQA-3k~\cite{agiqa3k} dataset due to its size. During training, the loss for each image is computed as the average loss over 3 random spatial crops. For inference, we average the predictions over 15 spatial crops per image to obtain a more stable and accurate assessment. All experiments were conducted using standard deep learning frameworks. Training was performed on NVIDIA RTX 4090 GPUs.

\begin{table}[htbp]
  \centering
  \vspace{-1em}
  \caption{Performance comparison on the AGIQA-1K dataset. Best results are highlighted in \textbf{bold}. Red values indicate the absolute percentage point improvement (\textit{i.e.}, of our method over the previous best result (underlined).}
  \label{tab:agiqa1k_results_abs}
  \vspace{-12pt}
  \begin{tabular}{l ccc}
    \toprule
    Method              & SRCC$\uparrow$            & PLCC$\uparrow$            & Main Score$\uparrow$      \\
    \midrule
    ResNet50~\cite{agiqa1k}          & 0.6365          & 0.7323          & 0.6844          \\
    StairIQA~\cite{agiqa1k}          & 0.5504          & 0.6088          & 0.5796          \\
    MGQA~\cite{agiqa1k}               & 0.6011          & 0.6760          & 0.6386          \\
    TIER~\cite{TIER}                & 0.8266          & 0.8297          & 0.8282          \\
    IP-IQA~\cite{BTPAGIQA}        & 0.8401          & 0.8922          & 0.8662          \\
    MoE-AGIQA-v1~\cite{MoE-AGIQA}       & \underline{0.8530} & 0.8877          & 0.8704          \\
    MoE-AGIQA-v2~\cite{MoE-AGIQA}       & 0.8501          & \underline{0.8922} & \underline{0.8712} \\
    \midrule
    SC-AGIQA (Ours)     & \textbf{0.9051} & \textbf{0.9558} & \textbf{0.9305} \\
    % --- Absolute Improvement Row ---
                        & \textcolor{red}{+5.21\%} & \textcolor{red}{+6.36\%} & \textcolor{red}{+5.93\%} \\
    \bottomrule
  \vspace{-2em}
  \end{tabular}
\end{table}
\begin{table}[htbp]
  \centering
  \caption{Performance comparison on the AGIQA-3K dataset. Best results are highlighted in \textbf{bold}. Red values indicate the absolute percentage point improvement (\textit{i.e.}, of our method over the previous best result (underlined).}
  \label{tab:agiqa3k_results_abs}
  \vspace{-12pt}
  \begin{tabular}{l ccc}
    \toprule
    Method              & SRCC$\uparrow$            & PLCC$\uparrow$            & Main Score$\uparrow$      \\
    \midrule
    DBCNN~\cite{agiqa3k}               & 0.8207          & 0.8759          & 0.8483          \\
    CLIPIQA~\cite{agiqa3k}             & 0.8426          & 0.8053          & 0.8240          \\
    CNNIQA~\cite{agiqa3k}               & 0.7478          & 0.8469          & 0.7974          \\
    HyperIQA~\cite{su2020blindly}   &   0.8526   & 0.8975    &  0.8751      \\
    TIER~\cite{TIER}               & 0.8251          & 0.8821          & 0.8536          \\
    IP-IQA~\cite{BTPAGIQA}        & 0.8634          & 0.9116          & 0.8875          \\
    IPCE~\cite{IPCE}              & 0.8841          & 0.9246          & 0.9044          \\
    MoE-AGIQA-v1~\cite{MoE-AGIQA}       & 0.8758          & \underline{0.9294} & 0.9026          \\
    MoE-AGIQA-v2~\cite{MoE-AGIQA}       & 0.8746          & 0.9282          & 0.9014          \\
    MA-AGIQA~\cite{MA-AGIQA}       & \underline{0.8939} & 0.9273          & \underline{0.9106} \\
    \midrule
    SC-AGIQA(Ours)      & \textbf{0.9070} & \textbf{0.9361} & \textbf{0.9216} \\
    % --- Absolute Improvement Row ---
                        & \textcolor{red}{+1.31\%} & \textcolor{red}{+0.67\%} & \textcolor{red}{+1.10\%} \\
    \bottomrule
    \vspace{-2em}
  \end{tabular}
\end{table}
\begin{table}[htbp]
  \centering
  \caption{Performance comparison on the AIGCIQA2023 dataset. Best results are highlighted in \textbf{bold}. Red values indicate the absolute percentage point improvement of our method over the previous best result (underlined).}
  \label{tab:aigciqa2023_results_abs}
  \vspace{-12pt}  
  \begin{tabular}{l ccc}
    \toprule
    Method              & SRCC$\uparrow$            & PLCC$\uparrow$            & Main Score$\uparrow$      \\
    \midrule
    CNNIQA~\cite{aigciqa2023}             & 0.7160          & 0.7937          & 0.7549          \\
    VGG16~\cite{aigciqa2023}              & 0.7961          & 0.7973          & 0.7967          \\
    VGG19~\cite{aigciqa2023}              & 0.7733          & 0.8402          & 0.8068          \\
    ResNet18~\cite{aigciqa2023}           & 0.7583          & 0.7763          & 0.7673          \\
    ResNet34~\cite{aigciqa2023}           & 0.7229          & 0.7578          & 0.7404          \\
    TIER~\cite{TIER}                 & 0.8194          & 0.8359          & 0.8277          \\
    HyperIQA~\cite{su2020blindly}   &  0.8357  &   0.8504    &  0.8431      \\
    IPCE~\cite{IPCE}              & 0.8640        & 0.8788           & 0.8714          \\
    MoE-AGIQA-v1~\cite{MoE-AGIQA}       & 0.8729          & 0.8860          & 0.8795          \\
    MoE-AGIQA-v2~\cite{MoE-AGIQA}       & \underline{0.8751} & \underline{0.8904} & \underline{0.8828} \\
    \midrule
    SC-AGIQA(Ours)      & \textbf{0.9113}      & \textbf{0.9158}      & \textbf{0.9136}   \\
    % --- Absolute Improvement Row ---
                        & \textcolor{red}{+3.62\%} & \textcolor{red}{+2.54\%} & \textcolor{red}{+3.08\%} \\
    \bottomrule
    \vspace{-3em}
  \end{tabular}
\end{table}
\subsection{Comparison with State-of-the-Art Methods}
\label{subsec:sota_comparison}
\begin{figure*}[ht] % Changed [ht] to [t] for figure* correctness
   \vspace{-1em}
  \centering
  \includegraphics[width=\textwidth]{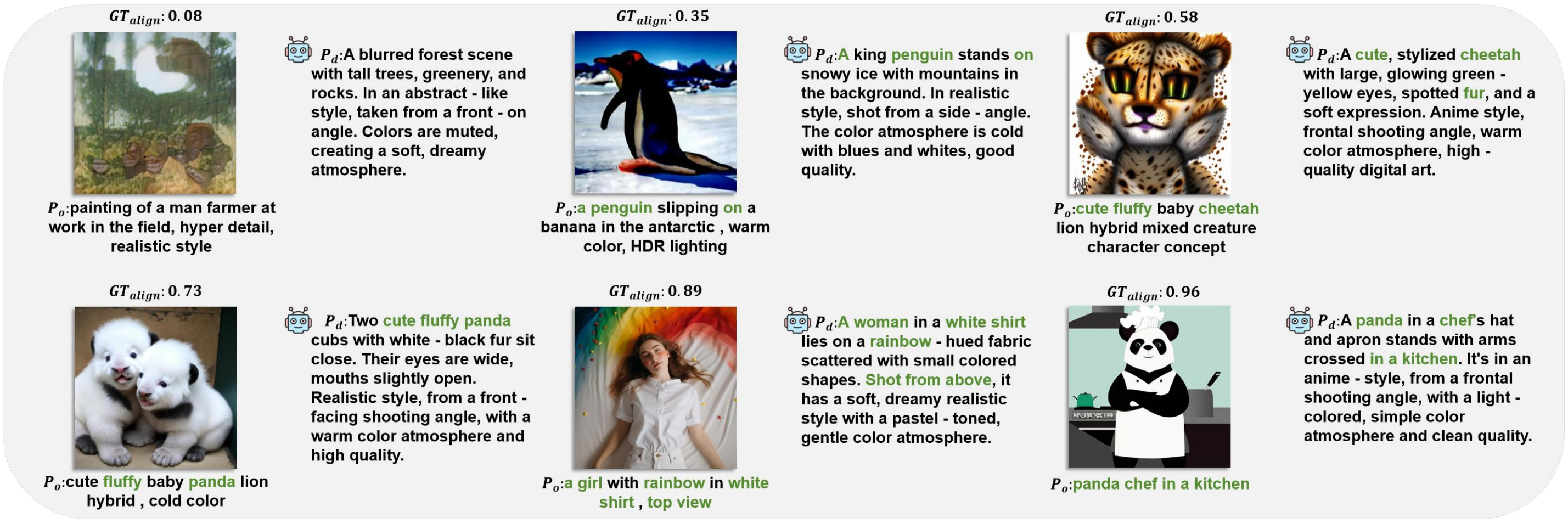}
  \caption{The $P_o$ and $P_d$ of images corresponding to different text-image alignment scores. As the text - image alignment scores increase, the semantic consistency between $P_o$ and $P_d$ becomes more pronounced. (green)}
  \label{fig:fig3}
  \vspace{-12pt}
\end{figure*}

To validate our proposed SC-AGIQA framework, we conducted comparisons against state-of-the-art (SOTA) methods on three AGIQA benchmarks. Performance was evaluated using the SRCC, PLCC, and Main Score, with results summarized in Tables~\ref{tab:agiqa1k_results_abs}, \ref{tab:agiqa3k_results_abs}, and \ref{tab:aigciqa2023_results_abs}.
On the AGIQA-1k dataset (Table~\ref{tab:agiqa1k_results_abs}), SC-AGIQA demonstrates superiority over previous methods. It achieves SRCC, PLCC, and Main Score of 0.9051, 0.9558, and 0.9305, respectively. These results represent absolute improvements of 5.21, 6.36, and 5.93 percentage points over the previously best-reported SRCC (MoE-AGIQA-v1) and PLCC/Main Score (MoE-AGIQA-v2), establishing a new benchmark for this dataset.

Similarly, on the AGIQA-3k dataset (Table~\ref{tab:agiqa3k_results_abs}), SC-AGIQA outperforms prior works, including the MA-AGIQA and MoE-AGIQA baselines. We obtain scores of 0.9070 (SRCC), 0.9361 (PLCC), and 0.9216 (Main Score), surpassing the previous SOTA (MA-AGIQA for SRCC/Main Score, MoE-AGIQA-v1 for PLCC) by 1.31, 0.67, and 1.10 percentage points, respectively. This highlights the robustness of our approach across different generative models and content types.

The performance advantage is confirmed on the AIGCIQA2023 benchmark (Table~\ref{tab:aigciqa2023_results_abs}), which features images from recent generative models. SC-AGIQA achieves scores of 0.9113 (SRCC), 0.9158 (PLCC), and 0.9136 (Main Score). This yields gains of 3.62, 2.54, and 3.08 percentage points over the competitive MoE-AGIQA-v2 across all metrics, demonstrating our method's capability in handling the nuances of modern AGI outputs.
\vspace{-1em} 
\subsection{Ablation Study}
To evaluate the contributions of the core components within our SC-AGIQA framework, we conducted ablation studies on the AGIQA-1K and AIGCIQA2023 datasets, with results presented in Table~\ref{tab:ablation_study_datasets}. We analyzed the impact of the Text-assisted Semantic Alignment Module (resulting in $SCI$), the Adaptive Quality-Aware Feature Pooling (AQAFP) within the FFDPM, and the integration of Human Preference (HP) signals.

We examined the role of semantic consistency by ablating the $SCI$ pathway. This removal caused a substantial performance drop on AGIQA-1k (e.g., SRCC decreasing from 0.9051 to 0.8649 when compared to the 'AQAFP+HP' setting), indicating the importance of assessing text-image alignment for this dataset. We posit this is because AGIQA-1k contains images from earlier models with more variable prompt adherence. In contrast, performance degradation in AIGCIQA2023 was less severe (SRCC dropping from 0.9113 to 0.8830), suggesting that the generally higher consistency of text and image in this more modern data set reduces the relative impact of explicit evaluation $SCI$ compared to aspects of visual quality.

The contribution of the AQAFP module was assessed by replacing it with average pooling within the FFDPM. Comparing configurations with and without AQAFP (e.g., 'SCI+HP' vs. the full model 'SCI+AQAFP+HP'), the inclusion of AQAFP improved performance across both datasets (e.g., on AIGCIQA2023, SRCC increased from 0.9016 to 0.9113), demonstrating the effectiveness of its adaptive spatial-channel weighting and integrated HVS priors in capturing perceptually relevant visual details beyond simple averaging.

Ablating the Human Preference (HP) signal integration—by removing the cross-attention fusion between visual features and preference-aware embeddings—also decreased performance on both benchmarks (e.g., comparing 'SCI+AQAFP' vs. the full model), confirming the benefit of injecting preference-related semantic context into the visual quality assessment.

The complete SC-AGIQA model, synergizing the MLLM-assisted semantic check ($SCI$), the perceptually enhanced pooling via AQAFP, and the HP integration, achieved the highest scores across all metrics on both datasets. This validates our design choices and underscores the necessity of jointly modeling nuanced semantic alignment and fine-grained visual quality for AGIQA.
\vspace{-1em}

\begin{figure*}[ht] % You might use [t] or [p] depending on placement needs
  \vspace{-1em}
  \centering
  \includegraphics[width=\textwidth]{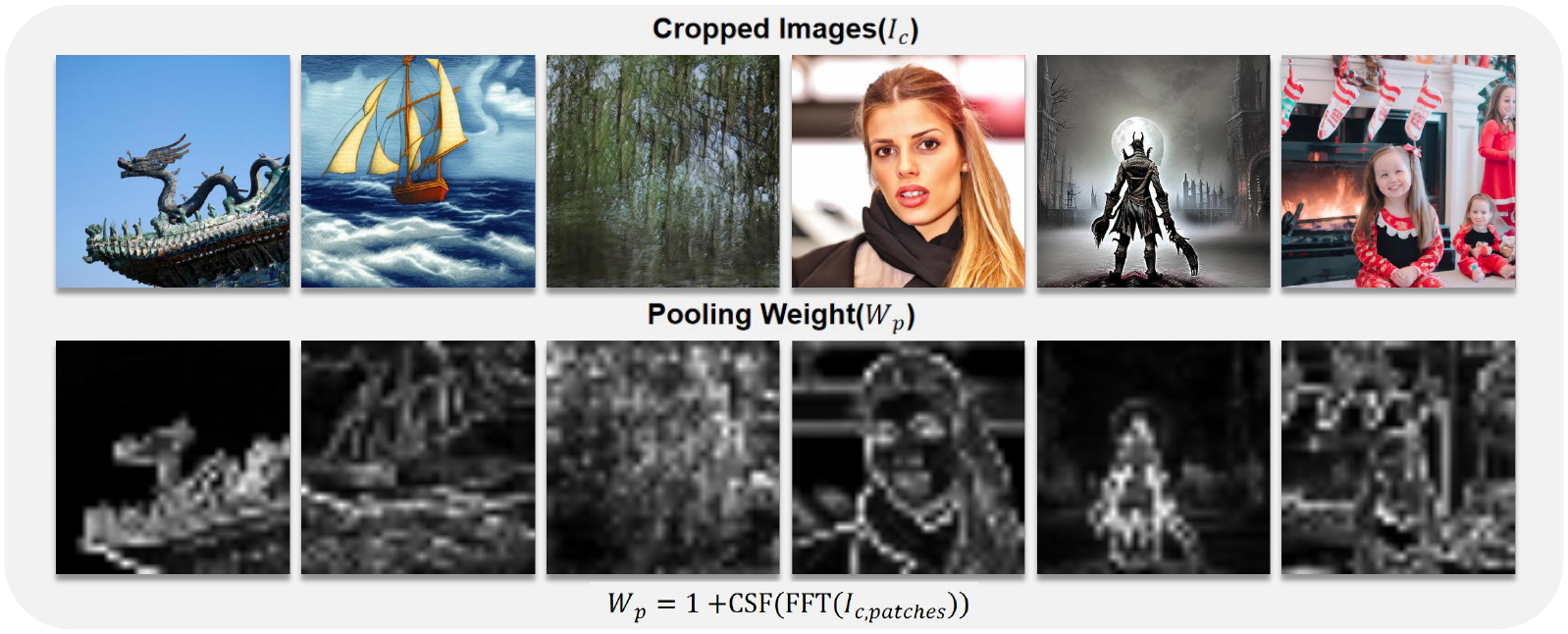} % Make sure the path/filename is correct
  \caption{HVS-based pooling weights ($W_p$) generated within the FFDPM's AQAFP module. Calculated as $W_p = 1 + \text{CSF}(\text{FFT}(I_{c,\text{patches}}))$, these weights (visualized below each cropped image $I_c$) prioritize perceptually salient frequency information for fine-grained visual quality assessment.}
  \label{fig:fig4} % Label for cross-referencing
  \vspace{-12pt}
\end{figure*}

\begin{table}[htbp] % htbp are placement specifiers: here, top, bottom, page
    \centering      % Center the table horizontally
    \caption{Ablation study of components (SCI, AQAFP, HP) on AGIQA-1K and AIGCIQA2023.} % Updated caption
    \label{tab:ablation_study_datasets} % Updated label for cross-referencing
    \vspace{-1em}
    \begin{tabular}{ccccccc} % 3 component cols + 2*2 metric cols = 7 cols
    \toprule % Top rule
    
    % --- Header Section ---
    % Row 1: Component headers (multirow) and Dataset headers (multicolumn)
    \multirow{2}{*}{SCI} & \multicolumn{2}{c}{VQI} & \multicolumn{2}{c}{AGIQA1K} & \multicolumn{2}{c}{AIGCIQA2023} \\
    \cmidrule(lr){2-3}
    % Row 2: Rules under dataset names and Metric sub-headers
    \cmidrule(lr){4-5} \cmidrule(lr){6-7}
     & AQAFP & HP & SRCC$\uparrow$ & PLCC$\uparrow$ & SRCC$\uparrow$ & PLCC$\uparrow$ \\

    \midrule % Rule between header and data

    % --- Data Section ---
    % Only SCI
    $\checkmark$ &            &            & 0.8890 & 0.9430 & 0.8855 & 0.8923 \\ % Added AIGCIQA2023 data

    % Only AQAFP
                 &       $\checkmark$ &  & 0.8212 & 0.8670 & 0.8328 & 0.8765 \\ % Added AIGCIQA2023 data
    % Only HP
             &            & $\checkmark$ & 0.8512 & 0.8974 & 0.8928 & 0.9005 \\ % Added AIGCIQA2023 data
    % SCI + AQAFP
    $\checkmark$ & $\checkmark$ &            & 0.9012 & 0.9512 & 0.9070 & 0.9105 \\ % Added AIGCIQA2023 data
    % AQAFP + HP
                 & $\checkmark$ & $\checkmark$ & 0.8649 & 0.9002 & 0.8830 & 0.8881 \\ % Added AIGCIQA2023 data
    % SCI + HP
    $\checkmark$ &            & $\checkmark$ & 0.8992 & 0.9486 & 0.9016 & 0.9064 \\ % Added AIGCIQA2023 data

    % SCI + AQAFP + HP (All modules)
    $\checkmark$ & $\checkmark$ & $\checkmark$ & \textbf{0.9051} & \textbf{0.9558} & \textbf{0.9113} & \textbf{0.9158} \\ % Added AIGCIQA2023 data

    \bottomrule % Bottom rule
    \vspace{-3em}
    \end{tabular}
\end{table}

\subsection{Visualization}
To provide qualitative insights into the mechanisms underlying SC-AGIQA's performance, we present visualizations focusing on the key operations within TSAM and FFDPM.

\textbf{MLLM-Assisted Semantic Alignment.} Figure~\ref{fig:fig3} illustrates the core principle of the Text-assisted Semantic Alignment Module (TSAM). It displays pairs of original prompts ($P_o$) and MLLM-generated descriptive prompts ($P_d$) for images ($I_o$) exhibiting varying ground-truth text-image alignment scores ($GT_{align}$). As observed, when the $GT_{align}$ score increases, indicating better alignment between the original prompt and the generated image, the semantic consistency between $P_o$ and $P_d$ becomes markedly more pronounced. Key concepts and details specified in $P_o$ are more likely to be explicitly mentioned and accurately described in the MLLM-generated $P_d$ for well-aligned images (highlighted green text). This visualization supports our approach of leveraging the MLLM's comprehension to derive $P_d$ as a robust proxy for the image's generated content, thereby facilitating a more nuanced assessment of text-image consistency through $P_o$-$P_d$ comparison.

\textbf{HVS-Informed Perceptual Weighting.}
Figure~\ref{fig:fig4} illustrates the HVS-informed weighting employed by the AQAFP module within our FFDPM. It displays the perceptual pooling weights ($W_p = 1 + W_h$, derived from applying CSF to the FFT of image patches $I_c$ as per Eq.~\ref{eq:hvs_weights} and used in Eq.~\ref{eq:aqafp_weighted_pool_final}) as heatmaps below each patch. These weights clearly prioritize regions with mid-to-high frequencies, corresponding to visually important features like edges and textures, while down-weighting smoother, low-frequency areas or imperceptible noise. This selective emphasis, grounded in HVS principles (Section~\ref{ssec:ffdpm}), guides the AQAFP pooling to focus on perceptually relevant details, enhancing the model's ability to detect fine-grained distortions crucial for AGI quality assessment.

\subsection{Limitations and Future Work}
\textbf{Redundancy in Image-Derived Description.} While our proposed SC-AGIQA framework demonstrates strong performance, we identify a potential area for future refinement concerning the generation of the descriptive prompt ($P_d$). In the current design, $P_d$ is generated by the MLLM based solely on the input image ($I_o$) and a predefined directive, deliberately excluding the original prompt ($P_o$) to mitigate potential biases where the MLLM might hallucinate prompt details not present in the image. However, this independence can introduce redundancy into $P_d$. For instance, the MLLM might describe both content and aesthetic attributes (e.g., image style, composition quality) comprehensively, even if the user's original prompt $P_o$ only specified content requirements and lacked any aesthetic constraints. This potential mismatch means that $P_d$ could contain information irrelevant to the specific criteria laid out in $P_o$, possibly adding noise to the subsequent semantic consistency comparison used to derive $SCI$. A promising direction for future work, which was not explored in this study, involves investigating strategies to refine $P_d$ by incorporating information from $P_o$. This could involve conditioning the MLLM on both $I_o$ and $P_o$ during generation using carefully designed instructions, or developing a post-processing step where the initial $P_d$ is filtered or contextualized based on $P_o$ to emphasize aspects most relevant to the user's request. Such an approach might yield a more focused and potentially more accurate evaluation of text-image semantic alignment.

\section{Conclusion}
\label{sec:conclusion}
Evaluating the quality of AGIs presents significant challenges, particularly in ensuring semantic alignment with textual prompts and accurately perceiving fine-grained visual details. This paper introduced SC-AGIQA, a unified framework designed to address these critical issues. By leveraging a Multimodal Large Language Model within the Text-assisted Semantic Alignment Module to generate a descriptive prompt for robust consistency checking, and incorporating Human Visual System priors through frequency domain analysis in the Frequency-domain Fine-grained Degradation Perception Module, our approach effectively tackles semantic misalignment and enhances sensitivity to subtle visual distortions. Extensive experiments conducted on three benchmarks demonstrate the superiority of SC-AGIQA, establishing new state-of-the-art performance across all evaluated metrics. The results underscore the efficacy of synergizing advanced MLLM semantic understanding with perceptually-grounded visual feature extraction for comprehensive AGI quality assessment. While acknowledging the potential for refining descriptive prompt generation as discussed, we will further enhance context-aware comparison conditioned on the original prompt.

\bibliographystyle{ACM-Reference-Format}
%\bibliography{bibtex}

\begin{thebibliography}{10}
\providecommand{\url}[1]{#1}
\csname url@samestyle\endcsname
\providecommand{\newblock}{\relax}
\providecommand{\bibinfo}[2]{#2}
\providecommand{\BIBentrySTDinterwordspacing}{\spaceskip=0pt\relax}
\providecommand{\BIBentryALTinterwordstretchfactor}{4}
\providecommand{\BIBentryALTinterwordspacing}{\spaceskip=\fontdimen2\font plus
\BIBentryALTinterwordstretchfactor\fontdimen3\font minus \fontdimen4\font\relax}
\providecommand{\BIBforeignlanguage}[2]{{%
\expandafter\ifx\csname l@#1\endcsname\relax
\typeout{** WARNING: IEEEtran.bst: No hyphenation pattern has been}%
\typeout{** loaded for the language `#1'. Using the pattern for}%
\typeout{** the default language instead.}%
\else
\language=\csname l@#1\endcsname
\fi
#2}}
\providecommand{\BIBdecl}{\relax}
\BIBdecl

\bibitem{niqe}
A.~Mittal, R.~Soundararajan, and A.~C. Bovik, ``Making a “completely blind” image quality analyzer,'' \emph{IEEE Signal Processing Letters}, vol.~20, no.~3, pp. 209--212, 2012.

\bibitem{moorthy2011blind}
A.~K. Moorthy and A.~C. Bovik, ``Blind image quality assessment: From natural scene statistics to perceptual quality,'' \emph{IEEE Transactions on Image Processing}, vol.~20, no.~12, pp. 3350--3364, 2011.

\bibitem{mittal2012no}
A.~Mittal, A.~K. Moorthy, and A.~C. Bovik, ``No-reference image quality assessment in the spatial domain,'' \emph{IEEE Transactions on Image Processing}, vol.~21, no.~12, pp. 4695--4708, 2012.

\bibitem{zhang2015feature}
L.~Zhang, L.~Zhang, and A.~C. Bovik, ``A feature-enriched completely blind image quality evaluator,'' \emph{IEEE Transactions on Image Processing}, vol.~24, no.~12, pp. 5357--5370, 2015.

\bibitem{zhang2020blind}
W.~Zhang, K.~Ma, J.~Yan, D.~Deng, and Z.~Wang, ``Blind image quality assessment using a deep bilinear convolutional neural network,'' \emph{IEEE Transactions on Circuits and Systems for Video Technology}, vol.~30, no.~1, pp. 36--47, 2020.

\bibitem{agnolucci2024arniqa}
L.~Agnolucci, L.~Galteri, M.~Bertini, and A.~Del~Bimbo, ``Arniqa: Learning distortion manifold for image quality assessment,'' in \emph{Proceedings of the Winter Conference on Applications of Computer Vision}.\hskip 1em plus 0.5em minus 0.4em\relax IEEE, 2024, pp. 189--198.

\bibitem{liu2023multiscale}
M.~Liu, J.~Huang, D.~Zeng, X.~Ding, and J.~Paisley, ``A multiscale approach to deep blind image quality assessment,'' \emph{IEEE Transactions on Image Processing}, vol.~32, pp. 1656--1669, 2023.

\bibitem{chen2024topiq}
C.~Chen, J.~Mo, J.~Hou, H.~Wu, L.~Liao, W.~Sun, Q.~Yan, and W.~Lin, ``Topiq: A top-down approach from semantics to distortions for image quality assessment,'' \emph{IEEE Transactions on Image Processing}, vol.~33, pp. 2404--2418, 2024.

\bibitem{su2020blindly}
S.~Su, Q.~Yan, Y.~Zhu, C.~Zhang, X.~Ge, J.~Sun, and Y.~Zhang, ``Blindly assess image quality in the wild guided by a self-adaptive hyper network,'' in \emph{Proceedings of the IEEE/CVF Conference on Computer Vision and Pattern Recognition}, 2020, pp. 3667--3676.

\bibitem{qin2023data}
G.~Qin, R.~Hu, Y.~Liu, X.~Zheng, H.~Liu, X.~Li, and Y.~Zhang, ``Data-efficient image quality assessment with attention-panel decoder,'' in \emph{Proceedings of the AAAI Conference on Artificial Intelligence}, vol.~37, no.~2, 2023, pp. 2091--2100.

\bibitem{wang2023exploring}
J.~Wang, K.~C. Chan, and C.~C. Loy, ``Exploring clip for assessing the look and feel of images,'' in \emph{Proceedings of the AAAI Conference on Artificial Intelligence}, vol.~37, no.~2, 2023, pp. 2555--2563.

\bibitem{zhang2023blind}
W.~Zhang, G.~Zhai, Y.~Wei, X.~Yang, and K.~Ma, ``Blind image quality assessment via vision-language correspondence: A multitask learning perspective,'' in \emph{Proceedings of the IEEE/CVF Conference on Computer Vision and Pattern Recognition}, 2023, pp. 14\,071--14\,081.

\bibitem{zhang2024qbench}
Z.~Zhang, H.~Wu, E.~Zhang, G.~Zhai, and W.~Lin, ``Q-bench: A benchmark for multi-modal foundation models on low-level vision from single images to pairs,'' \emph{IEEE Transactions on Pattern Analysis and Machine Intelligence}, vol.~46, pp. 10\,404--10\,418, 2024.

\bibitem{wu2024qinstruct}
H.~Wu, Z.~Zhang, E.~Zhang, C.~Chen, L.~Liao, A.~Wang, K.~Xu, C.~Li, J.~Hou, G.~Zhai, G.~Xue, W.~Sun, Q.~Yan, and W.~Lin, ``Q-instruct: Improving low-level visual abilities for multi-modality foundation models,'' in \emph{Proceedings of the IEEE/CVF Conference on Computer Vision and Pattern Recognition}, 2024, pp. 25\,490--25\,500.

\bibitem{CLIPAGIQA}
Z.~Tang, Z.~Wang, B.~Peng, and J.~Dong, ``Clip-agiqa: Boosting the performance of ai-generated image quality assessment with clip,'' in \emph{Proceedings of the International Conference on Pattern Recognition}.\hskip 1em plus 0.5em minus 0.4em\relax Springer, 2025, pp. 48--61.

\bibitem{MoE-AGIQA}
Y.~Yang, J.~Fu, W.~Zhang, W.~Cao, L.~Liu, and H.~Peng, ``Moe-agiqa: Mixture-of-experts boosted visual perception-driven and semantic-aware quality assessment for ai-generated images,'' in \emph{Proceedings of the IEEE/CVF Conference on Computer Vision and Pattern Recognition Workshops}, 2024, pp. 6395--6404.

\bibitem{SFIQA}
Z.~Yu, F.~Guan, Y.~Lu, X.~Li, and Z.~Chen, ``Sf-iqa: Quality and similarity integration for ai generated image quality assessment,'' in \emph{Proceedings of the IEEE/CVF Conference on Computer Vision and Pattern Recognition}, 2024, pp. 6692--6701.

\bibitem{BTPAGIQA}
Z.~Tang, Z.~Wang, B.~Peng, and J.~Dong, ``Bringing textual prompt to ai-generated image quality assessment,'' in \emph{Proceedings of the IEEE International Conference on Multimedia and Expo}.\hskip 1em plus 0.5em minus 0.4em\relax IEEE, 2024, pp. 1--6.

\bibitem{TSP-MGS}
J.~Xia, L.~He, F.~Gao, K.~Zhang, L.~Li, and X.~Gao, ``Ai-generated image quality assessment based on task-specific prompt and multi-granularity similarity,'' \emph{arXiv preprint arXiv:2411.16087}, 2024.

\bibitem{MA-AGIQA}
P.~Wang, W.~Sun, Z.~Zhang, J.~Jia, Y.~Jiang, Z.~Zhang, X.~Min, and G.~Zhai, ``Large multi-modality model assisted ai-generated image quality assessment,'' in \emph{Proceedings of the ACM International Conference on Multimedia}, 2024, pp. 7803--7812.

\bibitem{TIER}
J.~Yuan, X.~Cao, J.~Che, Q.~Wang, S.~Liang, W.~Ren, J.~Lin, and X.~Cao, ``Tier: Text-image encoder-based regression for aigc image quality assessment,'' \emph{arXiv preprint arXiv:2401.03854}, 2024.

\bibitem{IPCE}
F.~Peng, H.~Fu, A.~Ming, C.~Wang, H.~Ma, S.~He, Z.~Dou, and S.~Chen, ``Aigc image quality assessment via image-prompt correspondence,'' in \emph{Proceedings of the IEEE/CVF Conference on Computer Vision and Pattern Recognition Workshops}, 2024, pp. 6432--6441.

\bibitem{agiqazongshu}
Y.~Tian, Y.~Liu, S.~Wang, and S.~Kwong, ``Quality assessment for text-to-image generation: A survey,'' \emph{IEEE MultiMedia}, 2025.

\bibitem{xu2023imagereward}
J.~Xu, X.~Liu, Y.~Wu, Y.~Tong, Q.~Li, M.~Ding, J.~Tang, and Y.~Dong, ``Imagereward: Learning and evaluating human preferences for text-to-image generation,'' in \emph{Proceedings of the Advances in Neural Information Processing Systems}, vol.~36, 2023, pp. 15\,903--15\,935.

\bibitem{pick}
Y.~Kirstain, A.~Polyak, U.~Singer, S.~Matiana, J.~Penna, and O.~Levy, ``Pick-a-pic: An open dataset of user preferences for text-to-image generation,'' in \emph{Proceedings of the Advances in Neural Information Processing Systems}, vol.~36, 2023, pp. 36\,652--36\,663.

\bibitem{clip}
A.~Radford, J.~W. Kim, C.~Hallacy, A.~Ramesh, G.~Goh, S.~Agarwal, G.~Sastry, A.~Askell, P.~Mishkin, J.~Clark, G.~Krueger, and I.~Sutskever, ``Learning transferable visual models from natural language supervision,'' in \emph{Proceedings of the International Conference on Machine Learning}.\hskip 1em plus 0.5em minus 0.4em\relax PMLR, 2021, pp. 8748--8763.

\bibitem{blip}
J.~Li, D.~Li, C.~Xiong, and S.~C.~H. Hoi, ``Blip: Bootstrapping language-image pre-training for unified vision-language understanding and generation,'' in \emph{Proceedings of the International Conference on Machine Learning}.\hskip 1em plus 0.5em minus 0.4em\relax PMLR, 2022, pp. 12\,888--12\,900.

\bibitem{vit}
A.~Dosovitskiy, L.~Beyer, A.~Kolesnikov, D.~Weissenborn, X.~Zhai, T.~Unterthiner, M.~Dehghani, M.~Minderer, G.~Heigold, S.~Gelly, J.~Uszkoreit, and N.~Houlsby, ``An image is worth 16x16 words: Transformers for image recognition at scale,'' \emph{arXiv preprint arxiv:2010:11929}, 2020.

\bibitem{imagenet1k}
J.~Deng, W.~Dong, R.~Socher, L.-J. Li, K.~Li, and L.~Fei-Fei, ``Imagenet: A large-scale hierarchical image database,'' in \emph{Proceedings of the IEEE/CVF Conference on Computer Vision and Pattern Recognition}.\hskip 1em plus 0.5em minus 0.4em\relax IEEE, 2009, pp. 248--255.

\bibitem{aigciqa2023}
J.~Wang, H.~Duan, J.~Liu, S.~Chen, X.~Min, and G.~Zhai, ``Aigciqa2023: A large-scale image quality assessment database for ai generated images: from the perspectives of quality, authenticity and correspondence,'' in \emph{Proceedings of the CAAI International Conference on Artificial Intelligence}.\hskip 1em plus 0.5em minus 0.4em\relax Springer Nature Singapore, 2023, pp. 46--57.

\bibitem{agiqa1k}
Z.~Zhang, C.~Li, W.~Sun, X.~Liu, X.~Min, and G.~Zhai, ``A perceptual quality assessment exploration for aigc images,'' in \emph{Proceedings of the IEEE International Conference on Multimedia and Expo Workshops}.\hskip 1em plus 0.5em minus 0.4em\relax IEEE, 2023, pp. 440--445.

\bibitem{agiqa3k}
C.~Li, Z.~Zhang, H.~Wu, W.~Sun, X.~Min, X.~Liu, G.~Zhai, and W.~Lin, ``Agiqa-3k: An open database for ai-generated image quality assessment,'' \emph{IEEE Transactions on Circuits and Systems for Video Technology}, vol.~34, no.~8, pp. 6833--6846, 2023.

\bibitem{agiqa20k}
C.~Li, T.~Kou, Y.~Gao, Y.~Cao, W.~Sun, Z.~Zhang, Y.~Zhou, Z.~Zhang, W.~Zhang, H.~Wu, X.~Liu, X.~Min, and G.~Zhai, ``Aigiqa-20k: A large database for ai-generated image quality assessment,'' in \emph{Proceedings of the IEEE/CVF Conference on Computer Vision and Pattern Recognition}, 2024, pp. 6327--6336.

\bibitem{laion}
C.~Schuhmann, R.~Beaumont, R.~Vencu, C.~Gordon, R.~Wightman, M.~Cherti, A.~Kolesnikov, B.~Schiele, S.~Gelly, and J.~Jitsev, ``Laion-5b: An open large-scale dataset for training next generation image-text models,'' in \emph{Proceedings of the Advances in Neural Information Processing Systems}, vol.~35, 2022, pp. 25\,278--25\,294.

\bibitem{coco}
T.-Y. Lin, M.~Maire, S.~Belongie, L.~Bourdev, R.~Girshick, J.~Hays, P.~Perona, D.~Ramanan, C.~L. Zitnick, and P.~Dollar, ``Microsoft coco: Common objects in context,'' in \emph{Proceedings of the European Conference on Computer Vision}.\hskip 1em plus 0.5em minus 0.4em\relax Springer International Publishing, 2014, pp. 740--755.

\bibitem{csf}
J.~Mannos and D.~J. Sakrison, ``The effects of a visual fidelity criterion of the encoding of images,'' \emph{IEEE Transactions on Information Theory}, vol.~20, no.~4, pp. 525--536, 1974.

\bibitem{diffusion}
J.~Ho, A.~Jain, and P.~Abbeel, ``Denoising diffusion probabilistic models,'' in \emph{Proceedings of the Advances in Neural Information Processing Systems}, vol.~33, 2020, pp. 6840--6851.

\bibitem{gan}
I.~J. Goodfellow, J.~Pouget-Abadie, M.~Mirza, B.~Xu, D.~Warde-Farley, S.~Ozair, A.~Courville, and Y.~Bengio, ``Generative adversarial nets,'' in \emph{Proceedings of the Advances in Neural Information Processing Systems}, vol.~27, 2014.

\end{thebibliography}

\end{document}